\documentclass[conference]{IEEEtran}
\IEEEoverridecommandlockouts

\usepackage{cite}
\usepackage{amsmath,amssymb,amsfonts}
\usepackage{algorithmic}
\usepackage{algorithm}
\usepackage{graphicx}
\usepackage{booktabs}
\usepackage{multirow}
\usepackage{tabularx}
\usepackage{array}
\usepackage{makecell}
\usepackage{textcomp}
\usepackage{xcolor}
\usepackage{hyperref}

\makeatletter
\let\ps@plain\ps@empty
\makeatother

\title{An LLM-Driven Workflow for Automated Process Control Strategy Generation and Tuning from Dynamic Process Models}

\author{
\begin{tabular}{cc}
\begin{tabular}[t]{@{}c@{}}
\textbf{Ari Luna Rueda}\\
\textit{Department of Chemical Engineering}\\
\textit{Imperial College London}\\
London, UK\\
ari.luna-rueda21@imperial.ac.uk
\end{tabular}
&
\begin{tabular}[t]{@{}c@{}}
\textbf{Eike Cramer}\\
\textit{Department of Chemical Engineering}\\
\textit{University College London}\\
London, UK\\
e.cramer@ucl.ac.uk
\end{tabular}
\\[1.2em]
\begin{tabular}[t]{@{}c@{}}
\textbf{Klaus Hellgardt}\\
\textit{Department of Chemical Engineering}\\
\textit{Imperial College London}\\
London, UK\\
k.hellgardt@imperial.ac.uk
\end{tabular}
&
\begin{tabular}[t]{@{}c@{}}
\textbf{Mehmet Mercang{\"o}z}\\
\textit{Department of Chemical Engineering}\\
\textit{Imperial College London}\\
London, UK\\
m.mercangoz@imperial.ac.uk
\end{tabular}
\end{tabular}
}

\begin{document}

\maketitle
\thispagestyle{empty}
\pagestyle{empty}

\begin{abstract}
We present a structured large-language-model-driven workflow for automated multi-variable control design from dynamic process models. The workflow decomposes the design task into constrained code-generation steps: plant-interface construction, normalization, manipulated-variable controlled-variable (MV-CV) pairing, controller specification, closed loop simulation, scenario generation, performance evaluation and Bayesian-optimization (BO) based tuning. Generated artifacts are executed and validated before downstream tasks proceed, and failed artifacts are repaired using validation feedback. The proposed approach is demonstrated on a nonlinear gas-preheater benchmark with coupled pressure and temperature dynamics. The generated workflow produces a physically consistent decentralized PI (proportional-integral) feedback-feedforward control structure and an executable tuning environment. Bayesian optimization  reduces the closed loop performance objective, which aggregates set-point tracking and disturbance-rejection errors for the controlled variables, by approximately \(26.5\%\) relative to the initial controller generated by the workflow, mainly through improved pressure-loop transient performance. This figure quantifies the automated tuning stage rather than a comparison against a manually designed controller. The results demonstrate the feasibility of using structured large-language-model-based code generation to construct executable control-design workflows, while also highlighting the need for broader validation on larger plantwide-control benchmarks.
\end{abstract}

\textbf{Keywords:}
Large language models, Agentic AI, Autonomous control design, Plantwide control, Bayesian optimization

\section{Introduction}
\label{sec:introduction}

The broader move toward autonomous industrial operation is creating a need for engineering workflows that can adapt control designs when process configurations, operating objectives, or unit interconnections change, as can occur in modular and reconfigurable process systems. In such settings, the control system cannot be treated as a fixed, one-time engineering artifact. Even if the overall process function is preserved, a new configuration may require controller retuning, revised loop pairings in a decentralized architecture, or modified feedforward compensation paths. Systematically automating these redesign steps is therefore an important requirement for self-adapting operation in reconfigurable process systems. This challenge sits between two broader automation trends: on the upstream side, simulation and modelling workflows are becoming increasingly capable of generating, identifying, or reassembling dynamic process models from available engineering information or modular building blocks \cite{MercangozCortinovisSchonborn2020}; on the downstream side, the automated generation of industrial control logic for implementation on plant automation systems is receiving growing attention. However, the intermediate engineering layer that converts a dynamic process model into a revised control structure, an executable closed-loop simulation environment, and a tuning workflow remains largely manual and expert-driven.  

This paper approaches this gap as a problem of AI-enabled engineering automation. The goal is to automate the engineering work that surrounds a given dynamic process model; normalizing the model, selecting loop pairings, specifying and implementing a controller, constructing evaluation scenarios, and tuning the resulting closed-loop. Many of these subtasks  are expressed as executable code, which makes large language models (LLMs) a suitable automation mechanism, while the decomposition of the overall task into meaningful subtasks can be guided by the established plantwide-control intuition of pre-trained LLMs.

Concretely, this paper presents a first step towards an agentic LLM-driven framework for self-adapting control redesign in modular process systems. Here, the term agentic is used in an implementation-specific sense: the high-level task sequence is predefined, but each step combines language-model-based code generation with execution, validation and repair, so that generated artifacts are checked and revised before downstream tasks proceed. The framework targets the bottleneck between upstream dynamic-model generation and downstream controller implementation. The workflow uses LLMs to generate executable code artifacts that normalize the process model, synthesize a decentralized feedback and feedforward control structure, and compile a parametrized closed-loop simulation block.  This closed-loop block then serves as an automated tuning environment: the workflow executes the step tests to estimate dominant settling times, constructs dynamic evaluation scenarios, and generates Bayesian-Optimization (BO) scripts to tune controller parameters. A gas preheating process case study is used as a proof-of-concept to demonstrate the end-to-end pipeline. 

\vspace{0.5em}
\noindent\textbf{Contributions.} The main contributions of this paper are:
\begin{enumerate}
    \item \textbf{A structured decomposition of automated control (re)design:} The control-design problem is decomposed into executable subtasks, from plant normalization and MV--CV pairing to closed-loop simulation, scenario construction and controller tuning.

    \item \textbf{An LLM-driven code-generation workflow for control design with validation and repair:} Each generated artifact is constrained by an explicit task specification, checked through syntax, import and export validation, and regenerated when validation fails.

    \item \textbf{Automated construction of a controller-tuning environment:} The workflow generates not only controller code, but also the simulation, evaluation and Bayesian-optimization routines required to tune the controller against a fixed dynamic scenario.

    \item \textbf{End-to-end demonstration on a nonlinear multi-variable process:} The framework is demonstrated on a coupled pressure--temperature gas preheater benchmark, where the generated PI/feedforward controller is successfully tuned without manual modification of the generated pipeline.
\end{enumerate}

The remainder of the paper is organized as follows. Section~\ref{sec:related-work} reviews related work on automation upstream and downstream of control design, plantwide control, automated tuning, and LLM-based control-design assistance. Section~\ref{sec:case-study} introduces the gas preheater case study and its input--output structure. Section~\ref{sec:Methodology} details the LLM-driven workflow for model normalization, controller generation, scenario construction, and tuning. Section~\ref{sec:Results_and_discussion} evaluates the generated controller and closed-loop performance. Finally, Section~\ref{sec:conclusion} summarizes the findings and outlines future work.

%section II: Related work (separated from introduction, serving as as a small  the literature review, as requested by the reviewer. All the citations kept.
\section{Related Work}
\label{sec:related-work}

\subsection{Automation upstream and downstream of control design}

On the modelling side, autonomous process-model identification approaches include the recurrent neural-network based workflow proposed by \cite{MercangozCortinovisSchonborn2020}, as well as more recent comparative assessments of machine learning and system-identification methods for process dynamics \cite{AhmedDelRioChanonaMercangoz2025}. On the implementation side, automated generation of industrial control logic and related artifacts for programmable logic controller (PLC) and distributed control systems (DCS) deployment has received growing attention, including classification frameworks for control-code generation, automated generation of simulation models for control-code testing, and more recent LLM-based approaches for PLC/DCS code and test-case generation \cite{BarthFay2013,KoziolekGruenerAshiwal2023,KoziolekAshiwalBandyopadhyayKR2024}. In parallel, recent work has begun to explore AI-assisted generation of control structures directly from process flow diagrams (PFDs), for example by formulating the prediction of control structures as a translation task from PFDs without control structures to PFDs with control structures \cite{HirtreiterSchulzeBalhornSchweidtmann2023}. These lines of work automate the stages before and after control design, but leave the design stage itself to human engineers.

\subsection{Plantwide control structure design}

The plantwide control literature has long recognized that control-system design involves structural decisions in addition to parameter tuning. Seminal work by \cite{Skogestad2000Plantwide} framed plantwide control as the search for self-optimizing control structures, while later studies developed dynamic-model- and optimization-based procedures for synthesizing plantwide control architectures and extending them to multiple forcing scenarios and operating regimes \cite{WangMcAvoy2001Plantwide,ChenMcAvoyZafiriou2004Plantwide}. Subsequent reviews have consolidated design techniques, benchmarks, and open challenges in plantwide control \cite{deGodoyGarcia2017Review,AlstadSkogestad2017SOCSurvey}. Despite this maturity, these contributions primarily provide methodologies for human engineers and do not offer an executable autonomous workflow that revises the control design directly from a given dynamic model.

\subsection{Automated controller tuning}
A related but distinct line of work addresses automatic controller tuning. In particular, BO  has recently been used for automated tuning of multi-variable controllers in process applications \cite{VanNiekerkLeRouxCraig2022,RichterLeRouxCraig2025}. These studies demonstrate that data-efficient optimization can tune feedback controllers effectively when the controller structure, parameterization, and evaluation scenario are already available. However, they generally assume that the control architecture and simulation setup have already been specified, and therefore do not address the earlier engineering steps of normalizing the plant model, selecting loop pairings, arranging feedforward paths, and constructing standardized dynamic evaluation scenarios.

\subsection{LLM-based control-design assistance}

LLMs are increasingly capable of producing structured code, interacting with external tools, and supporting multi-step workflows. Initial studies indicate LLMs' ability to handle control-oriented tasks \cite{vyas2025autonomous}. This creates an opportunity to move beyond using LLMs as sources of engineering suggestions and instead to embed them within code-generation and simulation-driven workflows in which intermediate artifacts are executed, checked, and revised. Ares-Milian et al. use LLMs to provide expert knowledge for input-output pairing and stage definition in decentralized controller auto-tuning \cite{ares2025automating}. These works show the growing relevance of LLM-based control-design assistance, but they do not focus on generating the complete executable process-control software stack from a dynamic process model, including plant wrapping, controller implementation, scenario generation, evaluation and Bayesian-optimization scripts. The present work targets exactly this gap.

\section{Case study: Gas Preheater Process}
\label{sec:case-study}

\begin{figure}[htbp]
    \centering
    \includegraphics[width=0.9\linewidth]{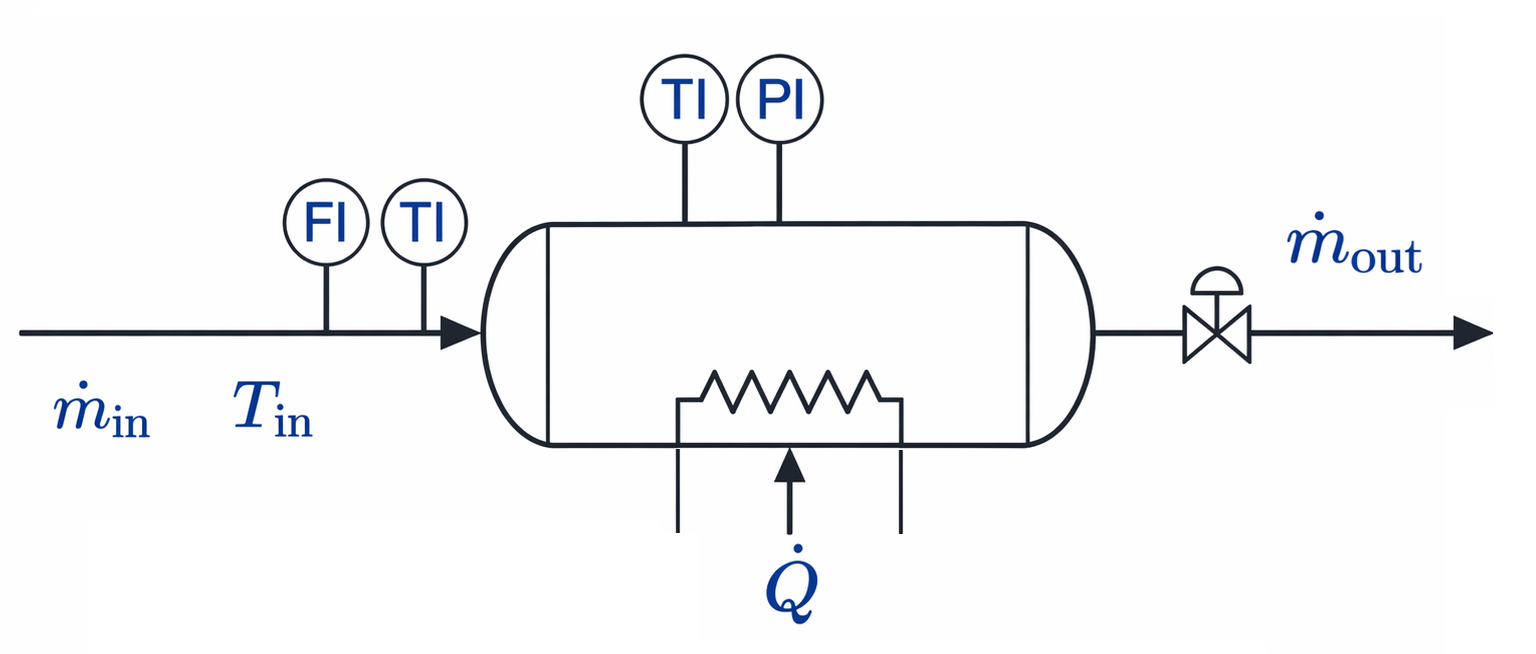} 
    \caption{Schematic of the gas-conditioning flow-header case study used for testing the control-design automation framework.}
    \label{fig:gas-tank-case-study}
\end{figure}

We use a fixed-volume flow header with an integrated electrical heating unit for gas conditioning as a case study, shown in Fig.~\ref{fig:gas-tank-case-study}. The system is a nonlinear two-input two-output process with coupled pressure and temperature dynamics. It contains an inlet stream, an outlet stream and an external heat input. The outlet mass flow rate, \(\dot{m}_{\mathrm{out}}\), and heat input, \(\dot{Q}\), are the manipulated variables, while the inlet mass flow rate, \(\dot{m}_{\mathrm{in}}\), and inlet temperature, \(T_{\mathrm{in}}\), are treated as measured disturbances. The controlled variables are the gas pressure and temperature inside the vessel. The benchmark therefore requires the controller to regulate both inventory and thermal dynamics while rejecting disturbances in the inlet conditions.

The manipulated variables are constrained by the actuator limits
\[
\dot{m}_{\mathrm{out}}\in[0,300]\ \mathrm{kg\,s^{-1}},
\qquad
\dot{Q}\in[-2.0\times10^8,\,2.0\times10^8]\ \mathrm{W}.
\]
In the source model, the measured disturbances are supplied using normalized source coordinates in the range \([0,1]\), which are mapped to physical values before the plant dynamics are evaluated.

The gas preheater has two states: the tank mass \(m\) and tank temperature \(T\). The governing equations are
\[
\begin{aligned}
\frac{dm}{dt} &= \dot{m}_{\mathrm{in}} - \dot{m}_{\mathrm{out}},
\\[0.8em]
\frac{dT}{dt}
&=
\frac{
\dot{m}_{\mathrm{in}}\left(c_p T_{\mathrm{in}} - c_v T\right)
-\dot{m}_{\mathrm{out}} R_s T
+\dot{Q}
}{
m c_v
},
\\[0.8em]
P &= \frac{m R_s T}{V},
\end{aligned}
\]
where \(R_s=R/M\) is the specific gas constant, \(V\) is the tank volume, and \(c_p\) and \(c_v\) are the constant-pressure and constant-volume heat capacities, respectively. The coupling between the manipulated variables and controlled variables is evident from these equations. The outlet flow rate directly affects the tank mass and therefore the pressure, while the heat input directly affects the temperature and indirectly affects the pressure through the ideal-gas relationship.

The model is implemented in the source file using two callable functions. The function \texttt{gas\_tank\_ODES} evaluates the state derivatives, while \texttt{gas\_tank\_outputs} converts the internal states into the physical output variables, pressure and temperature. The internal states are stored using a multiplicative scaling,
\[
x_s =
\begin{bmatrix}
m/S_{\mathrm{mass}} \\
T/S_{\mathrm{temp}}
\end{bmatrix},
\]
with \(S_{\mathrm{mass}}=503.6267\ \mathrm{kg}\) and \(S_{\mathrm{temp}}=573.15\ \mathrm{K}\). The ordinary differential equation (ODE) function returns derivatives in the same scaled basis, which can be called by an ODE solver. A metadata dictionary, \texttt{MODEL\_METADATA}, defines the input--output ordering, variable roles, scaling basis, nominal conditions, and normalization rules used by the external wrapper and controller interface.

\section{Proposed Workflow}
\label{sec:Methodology}
\begin{figure*}[t!]
    \centering
    \includegraphics[width=0.71\textwidth]{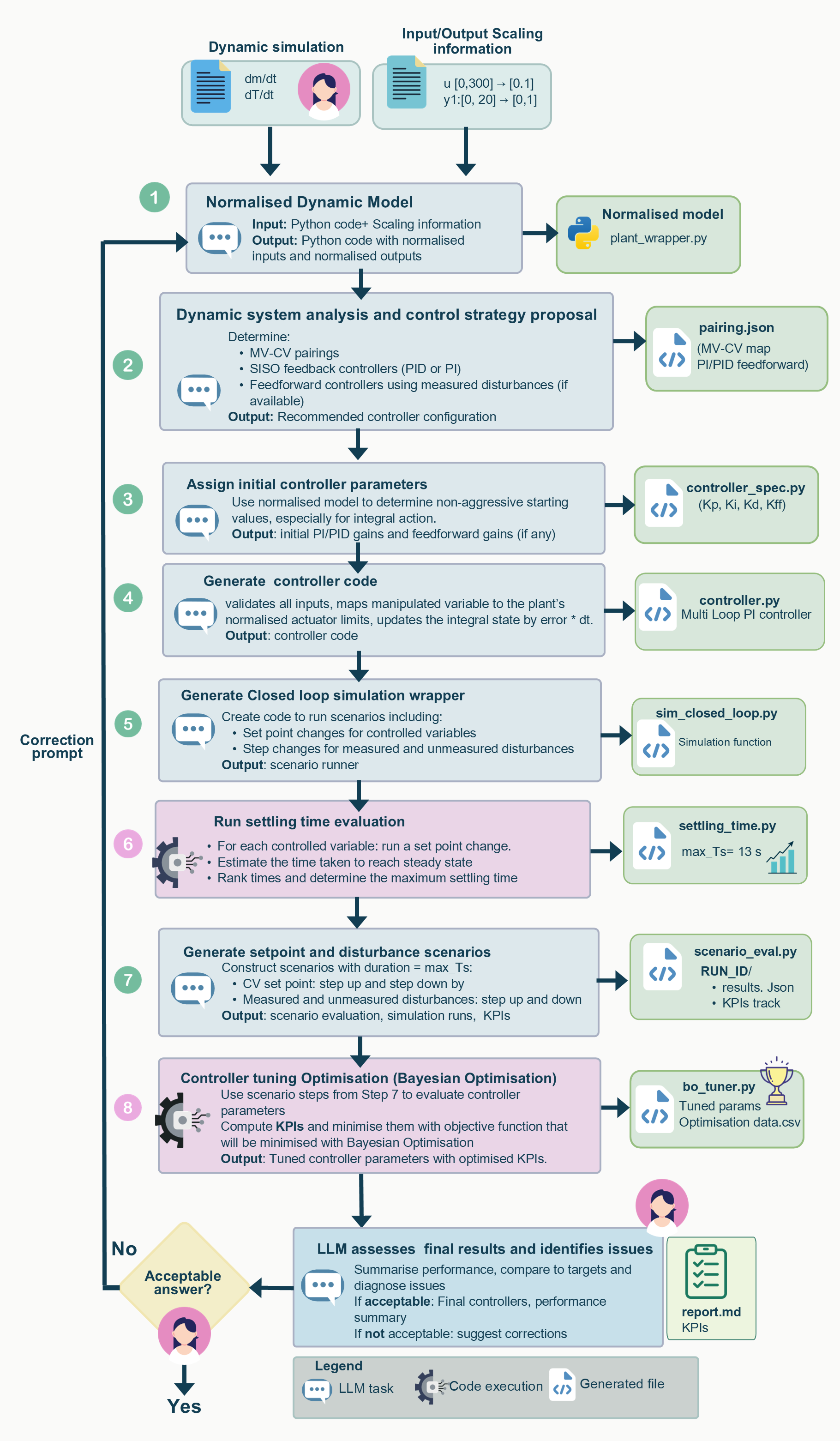}
    \caption{An agentic workflow for automated control system design. A user provides the problem description (in the form of controlled variable specifications) and an open-loop dynamic system model. The user receives the performance summary report and can be in the loop for performance assessment, which can also be carried out by a dedicated AI agent for a fully autonomous implementation. OpenAI GPT-5.4 was used as the LLM for implementing the framework.}
    \label{fig:framework}
\end{figure*}

The objective of the proposed workflow is to transform a given dynamic process model into an executable closed-loop control and tuning environment with minimal manual intervention, as illustrated in Fig.~\ref{fig:framework}. The workflow assumes that a source dynamic model and basic metadata are available, including state definitions, manipulated variables, controlled variables, measured disturbances, nominal conditions, scaling conventions and actuator bounds. 

The workflow is organized as a sequence of constrained code-generation tasks. Each task is defined by a machine-readable specification consisting of a target module, the upstream artifacts that must be read as context, the exact symbols the generated module must export, and a per-task generation budget. Because every task embeds the contents of previously generated modules as context, the agent never generates code in isolation, and later design decisions inherit and refine earlier ones. Each generation prompt imposes three layers of constraints: global rules (the source plant model must not be modified; output must be a single, importable, pure-Python module), task-specific requirements (exact exports, required behaviors, normalization rules and interface conventions), and the embedded upstream context. OpenAI GPT-5.4 was used as the code-generation model, called with system instructions that forbid markdown fences, explanations and the invention of additional files; tasks were generated with medium reasoning effort and per-task output budgets on the order of $10^{3}$--$10^{4}$ tokens.

\subsection{Validation and Repair Loop}

A key methodological feature of the workflow is the automatic validation and repair loop applied after each generation task. Once a file is generated, it is written through an atomic write operation to avoid partial file replacement. The generated file is then checked using task-specific validators. These checks include syntax validation, dynamic import validation and export validation against the symbols declared in the corresponding task specification. If validation succeeds, the task is marked as complete and the generated file becomes available as upstream context for subsequent tasks. If validation fails, the workflow constructs a repair prompt containing the original task instructions, the validation error message and the full contents of the failed file. This feedback mechanism does not guarantee semantic correctness, but it improves robustness relative to one-shot code generation and ensures that downstream tasks only receive artifacts that satisfy the declared interface-level checks.

\subsection{Functions of the generated modules}

\begin{figure}[tbp]
    \centering
    \includegraphics[width=0.5\linewidth]{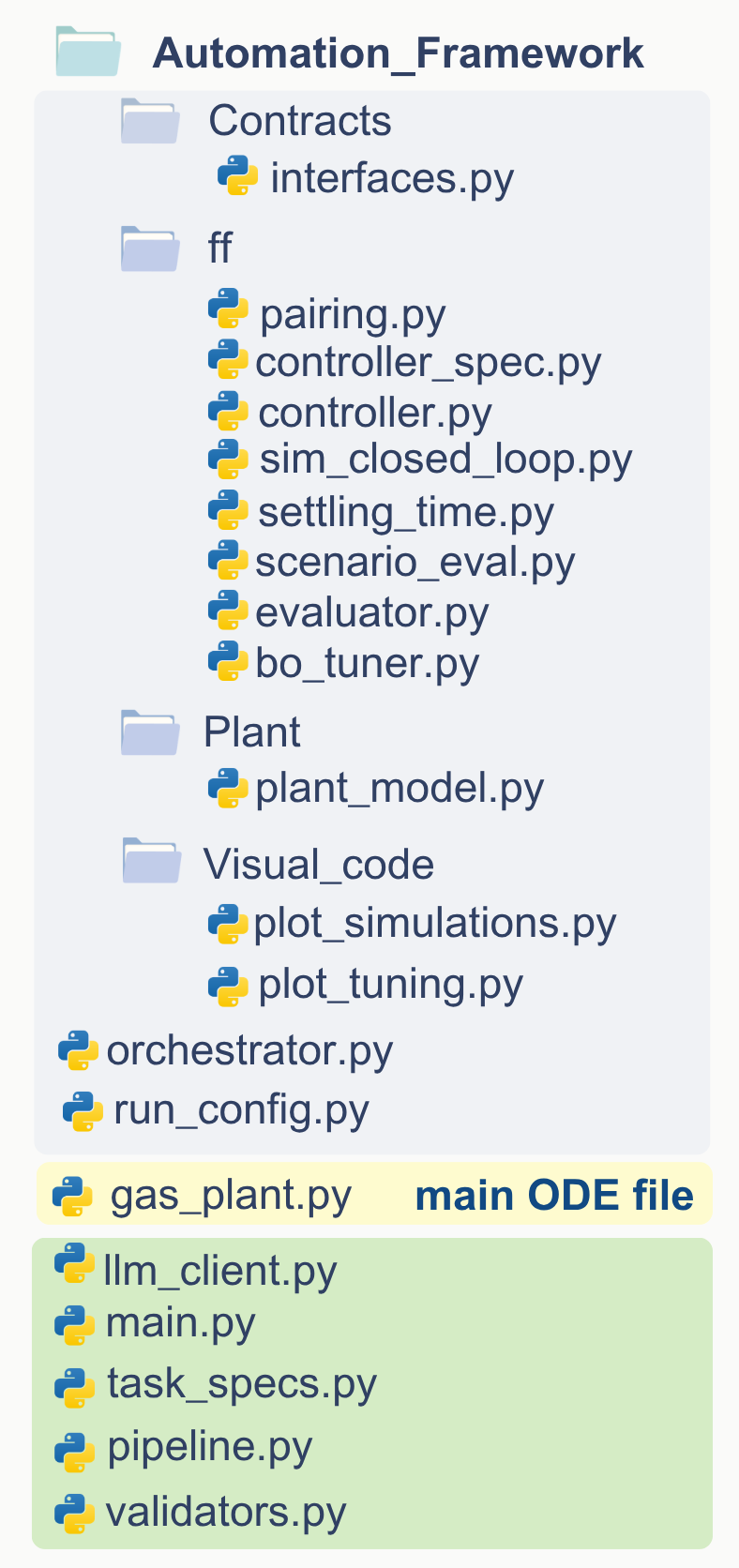}
    \caption{Folder structure used in the generated control-design workflow. Generated files are shown in blue, user-provided support code in green, and the original ODE model in yellow.}
    \label{fig:file_framework}
\end{figure}

The generated software stack comprises four functional layers (Fig.~\ref{fig:file_framework}); their roles are summarized below. 

\subsubsection{Plant contract and abstraction}
The first layer defines a formal intermediate representation of a dynamic plant: dimensions, callable right-hand-side and output functions, nominal operating values, actuator bounds and normalization metadata. An adapter module then loads the source model, extracts its metadata dictionary and exposes it through this validated representation. 
 
\subsubsection{Normalized plant wrapper}
The second layer wraps the validated plant in normalized coordinates, separating physical plant behavior from controller conventions. The plant state is kept in its native basis, while outputs, MVs and disturbances are normalized channel-by-channel according to the declared metadata. For a channel with nominal value $\bar v$ and span $s$, the nominal-centered map used for outputs and disturbances in this case study is
\begin{equation}
v_{\mathrm{n}} = \tfrac{1}{2} + \frac{v - \bar v}{2s},
\label{eq:norm-affine}
\end{equation}
which places the nominal operating point at $0.5$ and the range $\bar v \pm s$ onto $[0,1]$. The wrapper resolves the appropriate map for each channel from the metadata, constructs the inverse (denormalization) maps, and exposes normalized dynamics and output functions
\begin{equation}
\dot x = f\big(t, x, \mathcal{U}^{-1}(u_{\mathrm{n}}), \mathcal{D}^{-1}(d_{\mathrm{n}})\big),
\qquad
y_{\mathrm{n}} = \mathcal{N}_y\big(h(t, x, \cdot)\big),
\label{eq:wrapper}
\end{equation}
where $\mathcal{U}^{-1}$ and $\mathcal{D}^{-1}$ denormalize the commanded MVs and disturbances before evaluating the native model $f$ and output map $h$, and $\mathcal{N}_y$ normalizes the resulting outputs. 

\subsubsection{Control-structure synthesis}
The third layer determines the control structure and implements the controller. Pairing is decided from reasoning and quantitative evidence with open-loop probing of the wrapped plant near the nominal point. The pairing result is then compiled into a machine-readable controller specification, covering loop ordering, sign conventions, the parameter and controller-state layout, and fixed feedforward rules; for this case study, an inlet-mass compensation in the pressure loop and an energy-balance correction in the temperature loop. The controller declares conservative initial parameters $(K_{\mathrm{p}}, K_{\mathrm{i}}) = (1.0,\, 0.001)$ per loop together with global admissible ranges, which downstream tuning refines. 

\subsubsection{Simulation and tuning environment}
The fourth layer turns the plant and controller into a tuning environment. A single closed-loop simulation path applies set-point and disturbance events on a fixed control grid ($\Delta t_{\mathrm{c}} = 0.25$\,s in this study), calls the controller once per control interval and integrates the plant between control instants with four fixed fourth-order Runge--Kutta substeps. A settling-time module probes one CV at a time with a +0.1 normalized set-point step under the initial controller parameters and detects a departure-then-settle event. The design horizon $T_{\mathrm{block}}$ is the slowest measured settling time across CVs, and the workflow is configured to require a measured settling event for every controlled variable. The scenario generator uses $T_{\mathrm{block}}$ to construct a deterministic evaluation scenario: starting from nominal operation, each set-point and measured-disturbance channel is stepped up by $0.1$ (normalized), returned to nominal, stepped down and returned again, one channel at a time. The evaluator runs one closed-loop simulation of a candidate parameter vector $\theta$ on this fixed scenario and reduces the trajectories to a scalar objective. The objective function to minimize is defined as:
\begin{equation}
J(\theta) = \mathrm{MAE}_{\mathrm{n}}(\theta)\, \cdot\, \max_t \lvert e_{\mathrm{n}}(t;\theta)\rvert,
\label{eq:objective}
\end{equation}
the product of the mean and maximum absolute normalized tracking errors over the scenario, so that reducing sustained error is rewarded without permitting large transient excursions. The evaluator additionally computes tail-error, actuation-smoothness and saturation metrics for reporting, and returns a large penalty value ($10^{3}$) for simulation failures or guard violations, so the tuner can treat it as a total function. Finally, the Bayesian-optimization routine minimizes $J(\theta)$ over a local tuning box around the initial parameters, clipped to the controller's global bounds ($K_{\mathrm{p}}$ within $[0.5, 1.5]\times$ and $K_{\mathrm{i}}$ within $[0.25, 4]\times$ their initial values).  
 
\subsubsection{Orchestration and evidence}
A generated orchestrator executes the complete tuning workflow as a sequence of explicitly recorded stages: importing and re-validating all module exports, resolving the runtime configuration, building the fixed scenario, running the optimization, and replaying both the initial and the best controller on the identical scenario. Each stage writes a machine-readable summary, and the run produces a structured evidence record containing the resolved configuration, the pairing diagnostics, every controller evaluation with its parameters and metrics, incumbent-objective traces and the final comparison metrics. 

\begin{figure}[tbp]
    \centering
    \includegraphics[width=0.73\linewidth]{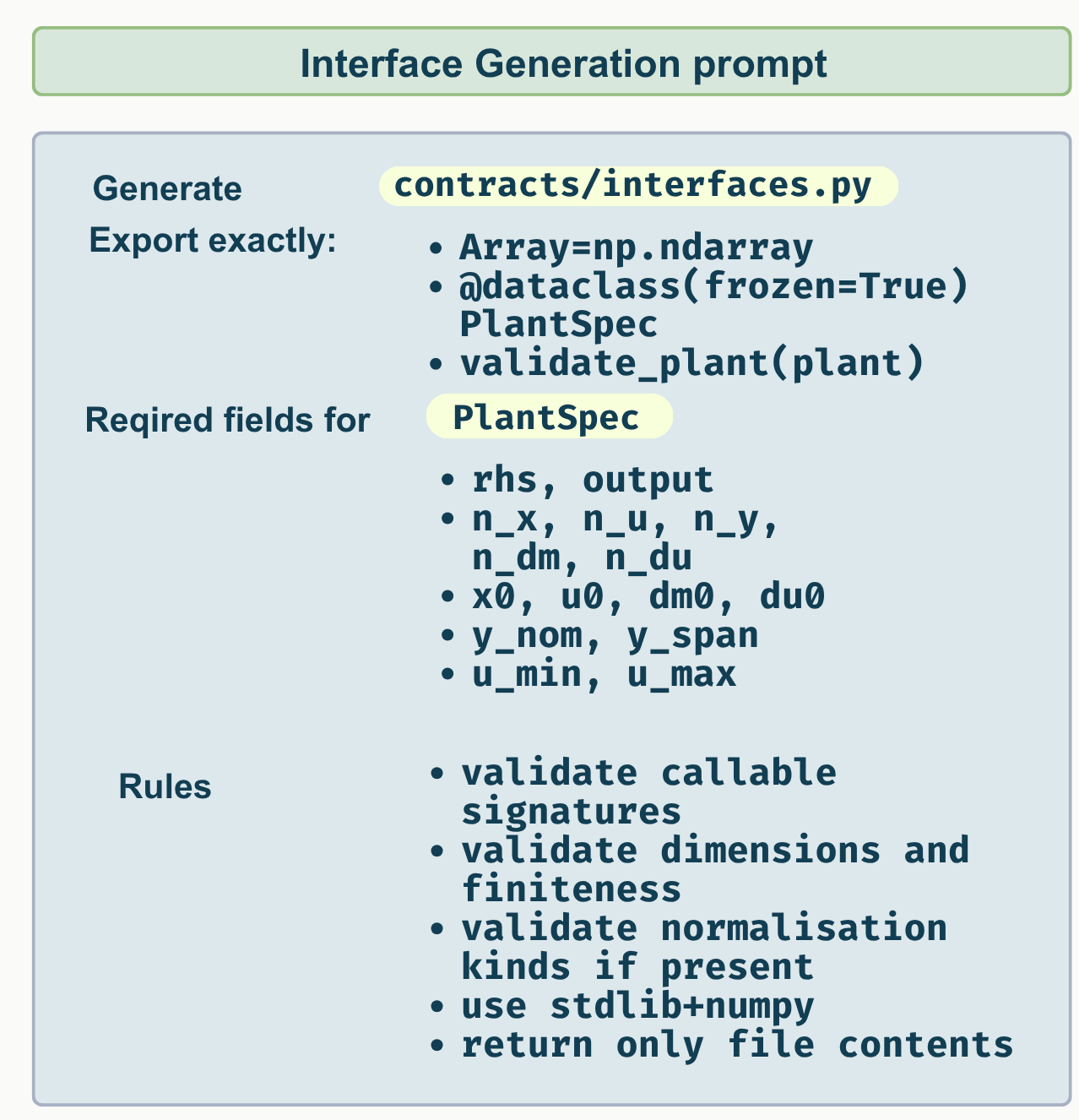}
    \caption{Prompt structure for plant-contract and interface generation.}
    \label{fig:interfaces_prompt}
\end{figure}

\begin{figure}[tbp]
    \centering
    \includegraphics[width=0.7\linewidth]{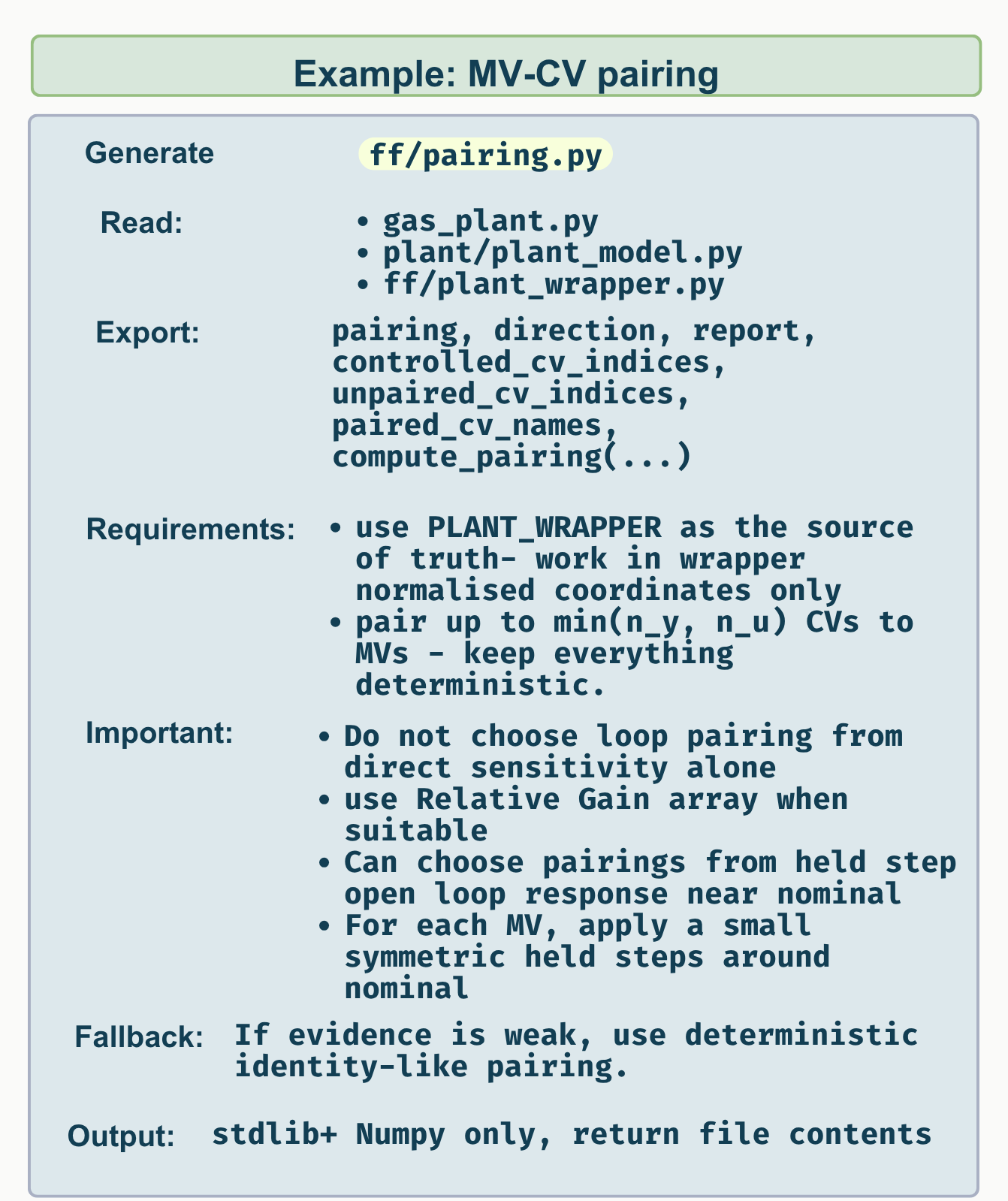}
    \caption{Pairing prompt structure.}
    \label{fig:pairing_prompt}
\end{figure}

\subsection{Representative generation prompts}
 
Figures~\ref{fig:interfaces_prompt} and~\ref{fig:pairing_prompt} show two of the structured prompts used in the pipeline. The interface-generation prompt (Fig.~\ref{fig:interfaces_prompt}) prescribes the plant data class, the expected metadata fields and the validation rules for dimensions, finiteness, callable signatures and normalization metadata, and therefore acts as the formal contract between the plant model and all downstream modules. The pairing prompt (Fig.~\ref{fig:pairing_prompt}) depends on the generated plant and wrapper modules as context, and requires the pairing logic to use the normalized wrapper as the source of truth, to evaluate deterministic held-step open-loop responses near the nominal operating point. 

These mechanisms contribute complementarily: task decomposition with embedded context bounds each generation step and surfaces interface errors where they arise; the contract and normalization layers keep downstream modules independent of source-model details such as scaling and variable ordering; the validation--repair loop converts generation failures into bounded retries, enabling unattended operation; the single shared simulation path ensures pairing, settling estimation and tuning observe identical dynamics; and freezing the scenario from the initial controller fixes the optimization landscape. A quantitative ablation of these mechanisms remains future work.

\section{Results and discussion}
\label{sec:Results_and_discussion}

\subsection{Generated control structure}

The generated controller uses a decentralized PI   (proportional-integral) architecture in normalized coordinates, with pressure paired with outlet mass flow rate $\dot{m}_{\mathrm{out}}$ and temperature paired with heat input $\dot Q$. In the comparisons below, the baseline controller refers to the initial controller produced by the workflow before Bayesian optimization. This baseline uses the generated PI/feedforward control structure and the initial PI parameters proposed during controller generation. The optimized controller retains the same pairing, feedback/feedforward structure and simulation scenario; only the PI parameter vector is modified by Bayesian optimization. This pairing is physically consistent: $\dot{m}_{\mathrm{out}}$ directly affects tank inventory and pressure, while $\dot Q$ acts on the internal energy balance and temperature. The generated plant wrapper, pairing module, controller, closed-loop simulator, scenario generator, evaluator and BO  were executed in sequence without manual modification. This confirms that the LLM-generated files formed a compatible end-to-end control-design and tuning pipeline. 

\subsection{Closed-loop tuning performance}

\begin{figure}[tbp]
    \centering
    \includegraphics[width=\linewidth]{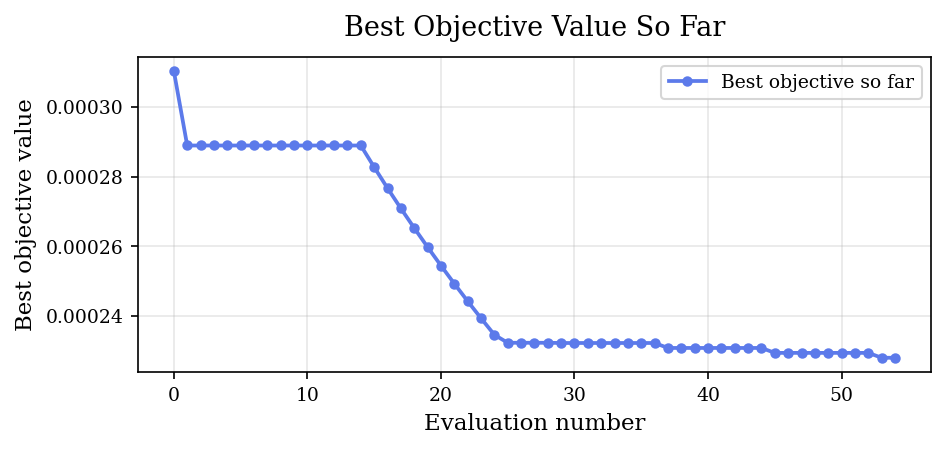} 
    \caption{Best-so-far objective value during Bayesian optimization.}
    \label{fig:objective-trajectory}
\end{figure}

\begin{figure}[htbp]
    \centering
    \includegraphics[width=0.95\linewidth]{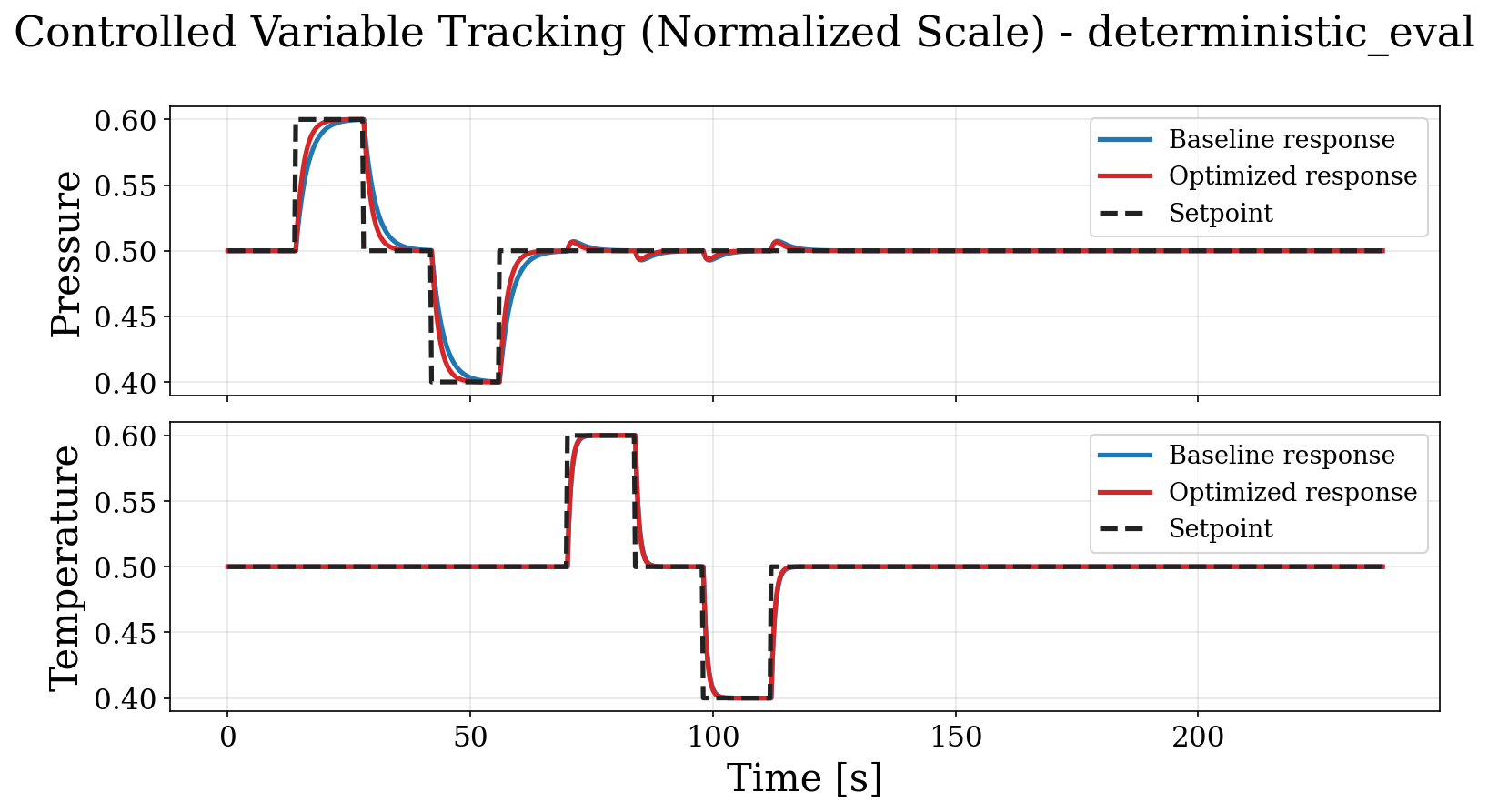} 
    \caption{Normalized controlled-variable tracking for the baseline and optimized controllers.}
    \label{fig:cv-tracking}
\end{figure}

\begin{figure}[hbtp]
    \centering
    \includegraphics[width=1.0\linewidth]{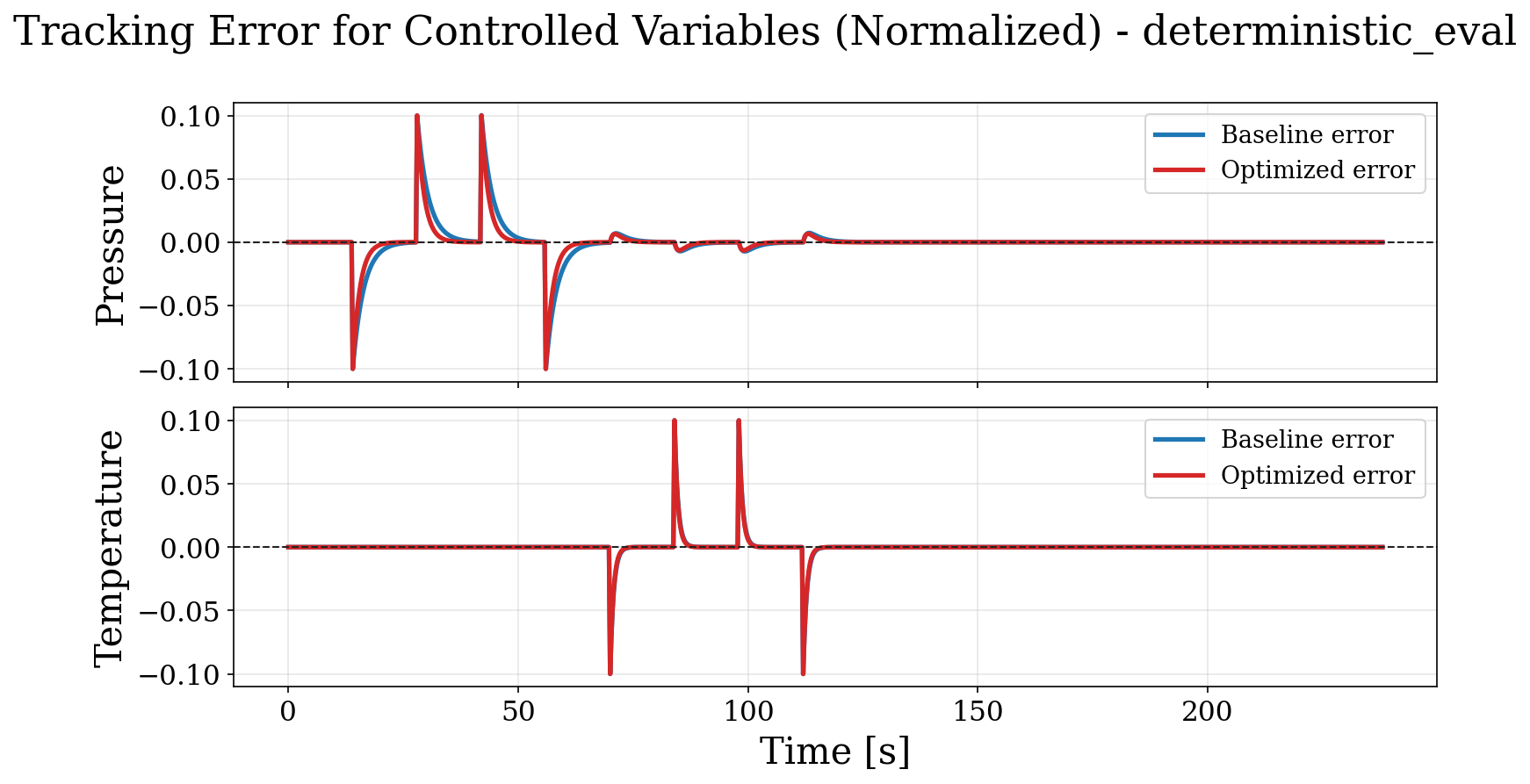} 
    \caption{Normalized tracking errors for the baseline and optimized controllers.}
    \label{fig:cv-error}
\end{figure}

The Bayesian-optimization objective used in this study reflects transient tracking performance for both controlled variables over the fixed scenario. The tuning workflow improved closed-loop performance. The raw objective decreased from $3.104\times10^{-4}$ to $2.280\times10^{-4}$, corresponding to a reduction of approximately $26.5\%$. The best-so-far trajectory in Fig.~\ref{fig:objective-trajectory} shows the run comprised the initial parameters, eight deterministic single-parameter seed points, six startup-mesh perturbations and 40 surrogate-guided expected-improvement iterations, for 55 evaluations in total, terminating at the configured iteration limit; all 55 evaluations completed without failures or penalties. The tuning loop is deterministic, so a rerun with the same configuration reproduces the same evaluation sequence: The main benefit is improved transient tracking rather than a large reduction in peak error. As shown in Figs.~\ref{fig:cv-tracking} and~\ref{fig:cv-error}, and summarized in Table~\ref{tab:error_metrics}, aggregate mean absolute error (MAE) and root mean squared error (RMSE) decrease by $26.37\%$ and $12.89\%$, respectively, whereas the maximum normalized error changes only marginally. The improvement is dominated by the pressure loop: pressure MAE decreases by $33.00\%$, compared with $5.65\%$ for temperature. The optimized parameters in Table~\ref{tab:controller_parameters} support this interpretation. The pressure-loop proportional gain increases to its upper local bound, while the integral gain decreases; the temperature-loop gains remain close to their baseline values.

\begin{table}[tbp]
\caption{Normalized error metrics for the baseline and optimized controllers. Lower values are better.}
\label{tab:error_metrics}
\centering
\footnotesize
\setlength{\tabcolsep}{3pt}
\renewcommand{\arraystretch}{1.05}
\resizebox{\columnwidth}{!}{%
\begin{tabular}{llccc}
\toprule
\textbf{Channel} & \textbf{Metric} & \textbf{Baseline} & \textbf{Optimized} & \textbf{\% Change} \\
\midrule
\multirow{6}{*}{Aggregate}
& MAE$_{\text{norm}}$        & 0.003095 & 0.002279 & -26.37 \\
& RMSE$_{\text{norm}}$       & 0.01249  & 0.01088  & -12.89 \\
& Final $|e|_{\text{norm}}$  & $6.561\times10^{-7}$ & $5.041\times10^{-7}$ & -23.16 \\
& Max $|e|_{\text{norm}}$    & 0.100287 & 0.100063 & -0.22 \\
& Tail MAE$_{\text{norm}}$   & $6.665\times10^{-7}$ & $5.115\times10^{-7}$ & -23.26 \\
& Tail RMSE$_{\text{norm}}$  & $9.001\times10^{-7}$ & $7.213\times10^{-7}$ & -19.86 \\
\midrule
\multirow{6}{*}{Pressure}
& MAE$_{\text{norm}}$        & 0.004691 & 0.003143 & -33.00 \\
& RMSE$_{\text{norm}}$       & 0.01500  & 0.01240  & -17.31 \\
& Final $|e|_{\text{norm}}$  & $6.025\times10^{-8}$ & $2.912\times10^{-9}$ & -95.17 \\
& Max $|e|_{\text{norm}}$    & 0.100287 & 0.100063 & -0.22 \\
& Tail MAE$_{\text{norm}}$   & $6.170\times10^{-8}$ & $2.941\times10^{-9}$ & -95.23 \\
& Tail RMSE$_{\text{norm}}$  & $6.171\times10^{-8}$ & $2.941\times10^{-9}$ & -95.23 \\
\midrule
\multirow{6}{*}{Temperature}
& MAE$_{\text{norm}}$        & 0.001500 & 0.001415 & -5.65 \\
& RMSE$_{\text{norm}}$       & 0.009333 & 0.009110 & -2.39 \\
& Final $|e|_{\text{norm}}$  & $1.252\times10^{-6}$ & $1.005\times10^{-6}$ & -19.70 \\
& Max $|e|_{\text{norm}}$    & 0.100057 & 0.100051 & -0.01 \\
& Tail MAE$_{\text{norm}}$   & $1.271\times10^{-6}$ & $1.020\times10^{-6}$ & -19.77 \\
& Tail RMSE$_{\text{norm}}$  & $1.271\times10^{-6}$ & $1.020\times10^{-6}$ & -19.77 \\
\bottomrule
\end{tabular}%
}
\end{table}

\begin{table}[b!]
\caption{Baseline and optimized controller parameters.}
\label{tab:controller_parameters}
\centering
\begin{tabular}{lcc}
\toprule
\textbf{Parameter} & \textbf{Baseline} & \textbf{Optimized} \\
\midrule
Pressure $K_p$ & 1.0 & 1.5 \\
Pressure $K_i$ & 0.001 & 0.00067 \\
Temperature $K_p$ & 1.0 & 1.092727 \\
Temperature $K_i$ & 0.001 & 0.001 \\
\bottomrule
\end{tabular}
\end{table}

\begin{figure}[htbp]
    \centering
    \includegraphics[width=\linewidth]{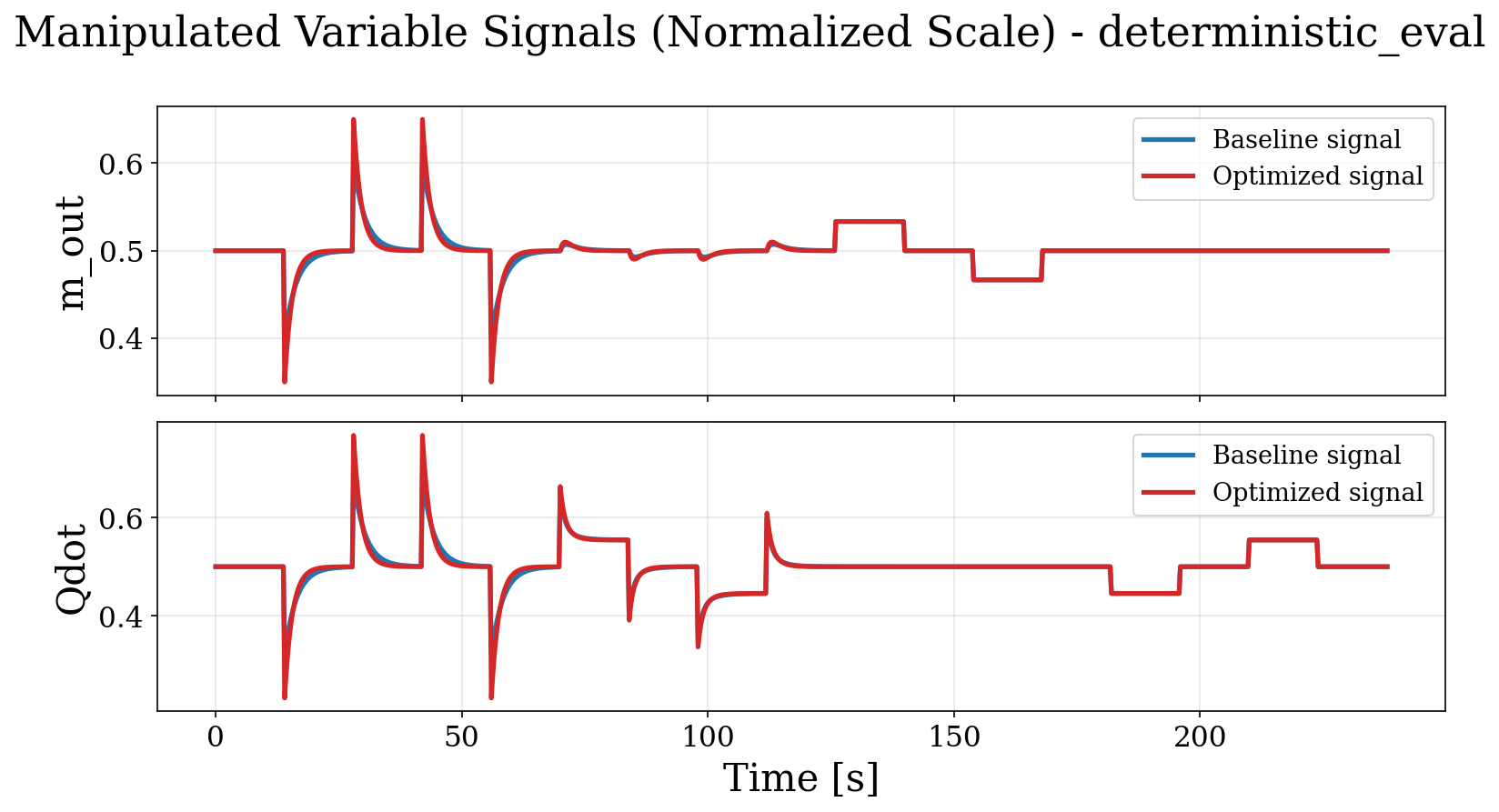} 
    \caption{Normalized manipulated-variable trajectories for $\dot{m}_{\mathrm{out}}$ and $\dot Q$.}
    \label{fig:mv-trajectories}
\end{figure}

\begin{figure}[htbp]
    \centering
    \includegraphics[width=0.95\linewidth]{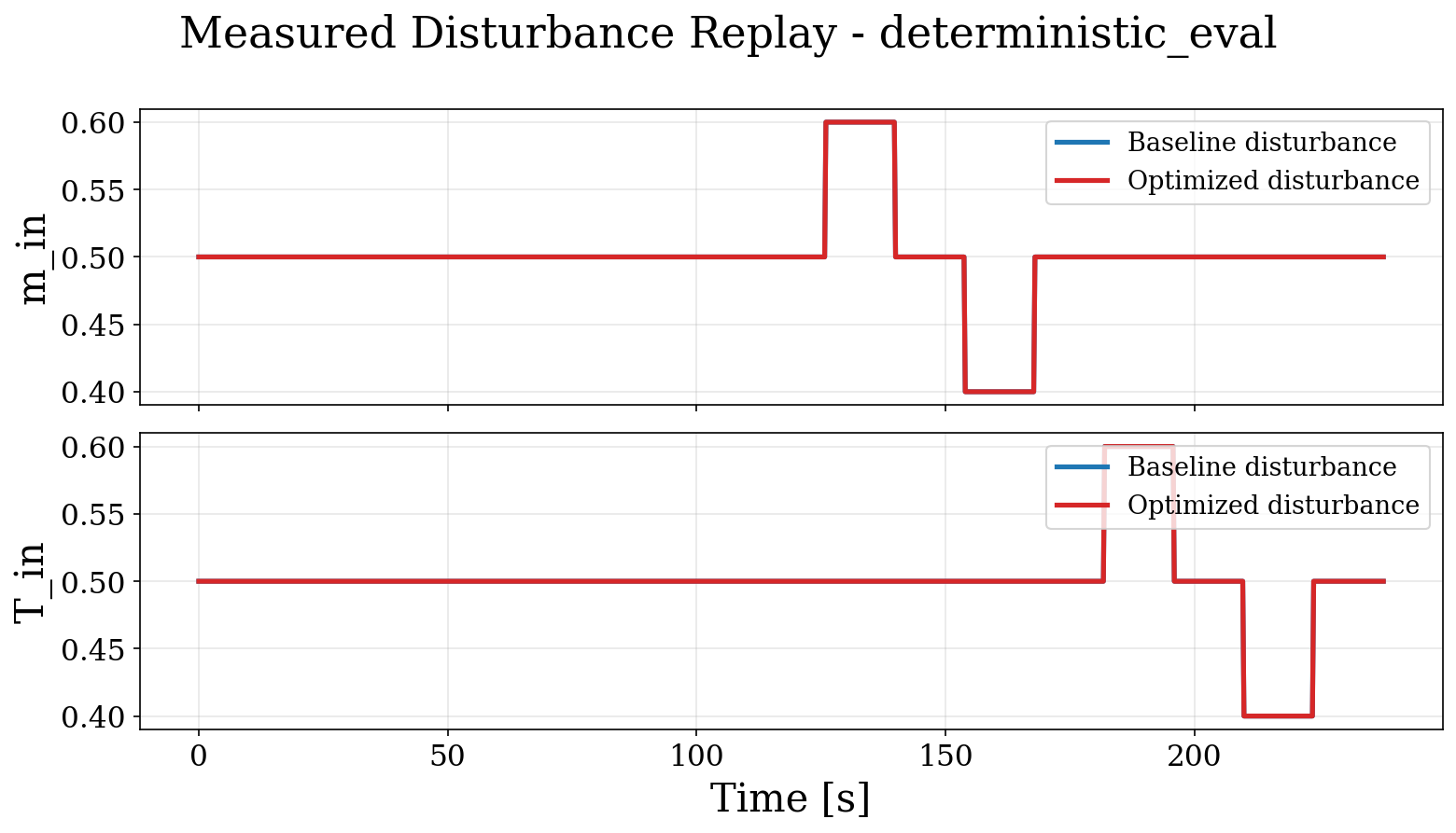} 
    \caption{Normalized measured-disturbance replay applied to both controllers.}
    \label{fig:measured-disturbances}
\end{figure}

The improved tracking is obtained through increased controller movement, as shown in Fig.~\ref{fig:mv-trajectories}. The optimized controller applies more assertive transient actuation, but both manipulated variables remain within their limits and the saturation metrics are zero. Since the generated controller does not yet include anti-windup, this remains a limitation for more demanding benchmarks. The measured-disturbance replay in Fig.~\ref{fig:measured-disturbances} confirms that the scenario includes both inlet-flow and inlet-temperature disturbances. Their impact is small because feedforward compensation removes the dominant effects and feedback corrects the residual error.

Overall, the optimized controller improves pressure regulation while preserving the strong temperature response, at the cost of increased but admissible actuator movement. Finally, the validation results should be interpreted as interface-level rather than control-theoretic guarantees. 

\subsection{Role of human users and user interaction}

Although the presented workflow is designed for autonomous operation, human interaction can occur at several points, as illustrated in Fig.~\ref{fig:framework}. A human user may provide the initial problem specification directly, in place of the upstream model-generation workflow, including the dynamic model, input--output definitions and control objectives. After controller generation and tuning, the user can inspect the final report, assess closed-loop performance and confirm the controller configuration. If performance is unsatisfactory, the user can provide additional feedback to guide subsequent LLM calls. The workflow can therefore serve as a human-in-the-loop design assistant, while also supporting more autonomous implementations in which assessment and feedback are done by AI agents.

\section{Conclusion and Outlook}
\label{sec:conclusion}

This work presented an LLM-driven workflow for generating, validating and tuning a closed-loop control pipeline from a dynamic process model. The workflow decomposes control redesign into a sequence of constrained code-generation tasks, including plant abstraction, normalization, MV--CV pairing, controller specification, closed-loop simulation, scenario construction, performance evaluation and Bayesian-optimization-based tuning. Intermediate validation and repair steps allow generated artifacts to be checked and revised before downstream execution.

The gas-preheater case study demonstrates the feasibility of the approach on a coupled nonlinear two-loop process. The generated controller selected a physically consistent decentralized PI/feedforward structure, and Bayesian-optimization-based tuning reduced the closed-loop objective by approximately \(26.5\%\) relative to the initial LLM-proposed parameterization, mainly through improved pressure-loop transients. These results show that structured LLM-driven code generation can support the construction of executable control-tuning workflows, however they are based on a single case study, a single LLM and a single end-to-end run, and should be interpreted as a feasibility demonstration rather than a statistical assessment of workflow robustness. Future work will focus on larger plantwide-control benchmarks, stronger validation of generated modules, incorporation of structured control-engineering knowledge into the generation context, and systematic quantification of generation failures, repair iterations and workflow robustness.

\bibliographystyle{IEEEtran}
\bibliography{bib/processes-v13-i09_20260127}
\end{document}